\documentclass[10pt,journal,compsoc]{IEEEtran}
\ifCLASSOPTIONcompsoc
  \usepackage[nocompress]{cite}
\else
  \usepackage{cite}
\fi
\usepackage{times}
\usepackage{threeparttable}
\usepackage{epsfig}
\usepackage{graphicx}
\usepackage{amsmath}
\usepackage{amssymb}
\usepackage{times}
\usepackage{epsfig}
\usepackage{graphicx}
\usepackage{amsmath}
\usepackage{amssymb}
\usepackage{multirow}
\usepackage{url}
\usepackage{bm} 
\usepackage{caption}
\usepackage{subfigure}
\usepackage{float} 
\usepackage{algpseudocode,algorithm}
\usepackage{gensymb}
\usepackage{booktabs}
\usepackage{xcolor}
\graphicspath{{figure/}}
\ifCLASSINFOpdf
\else
\fi
\begin{document}
%
\title{Contrastive Learning with Stronger Augmentations}
%
%
%
%

\author{Xiao Wang,~\IEEEmembership{Student~Member,~IEEE}
        and Guo-Jun Qi,~\IEEEmembership{Fellow,~IEEE}
\IEEEcompsocitemizethanks{\IEEEcompsocthanksitem X. Wang was with Department of Computer Science, Purdue University,West Lafayette, 47906, USA.
\IEEEcompsocthanksitem G-J. Qi was with the Futurewei Seattle Cloud Lab, Seattle, WA, 98006, USA. Email: guojunq@gmail.com.
}
\thanks{Manuscript received April, 2021; revised Dec, 2021; accepted xx,xx.(Corrresponding author: Guo-Jun Qi)}}

%
%

\markboth{IEEE Transactions on Pattern Analysis and Machine Intelligence,~Vol.~xx, No.~xx, April~2021}%
{Wang \MakeLowercase{\textit{et al.}}: Contrastive Learning with Stronger Augmentations}
%



\IEEEtitleabstractindextext{%
\begin{abstract}
Representation learning has significantly been developed with the advance of contrastive learning methods. Most of those methods have benefited from various data augmentations that are carefully designated to maintain their identities so that the images transformed from the same instance can still be retrieved. However, those carefully designed transformations limited us to further explore the novel patterns exposed by other transformations. Meanwhile, as found in our experiments, strong augmentations distorted the images' structures, resulting in difficult retrieval. Thus, we propose a general framework called Contrastive Learning with Stronger Augmentations~(CLSA) to complement current contrastive learning approaches. Here, the distribution divergence between the weakly and strongly augmented images over the representation bank is adopted to supervise the retrieval of strongly augmented queries from a pool of instances. Experiments on the ImageNet dataset and downstream datasets showed the information from the strongly augmented images can significantly boost the performance. For example, CLSA achieves top-1 accuracy of 76.2\% on ImageNet with a standard ResNet-50 architecture with a single-layer classifier fine-tuned, which is almost the same level as 76.5\% of supervised results. The code and pre-trained models are available in \url{https://github.com/maple-research-lab/CLSA}.
\end{abstract}

}

\maketitle

\IEEEdisplaynontitleabstractindextext

%
\IEEEpeerreviewmaketitle

\IEEEraisesectionheading{\section{Introduction}\label{sec:introduction}}

Deep neural networks have shown their sweeping successes in learning from large-scale labeled datasets like ImageNet~\cite{deng2009imagenet}. However, such successes hinge on the availability of a large number of labeled examples that are expensive to collect. To address this challenge, unsupervised visual representation learning and self-supervised learning have been extensively studied to learn feature representations without labels. Researchers aim to build a framework that can self-learn representations, which does not suffer from the request of many labels and can learn robust and general representations. Among them, the contrastive learning~\cite{hadsell2006dimensionality,misra2020self,chen2020improved,he2020momentum,caron2020unsupervised}, which is one formulation of the instance learning, shows significant potentials to close the performance gap with supervised methods. 

In instance learning~\cite{maron1998framework}, each image is being considered as an instance, and we wish to train the network so that the representations of different augmented views of the same instance are as close as possible to each other. Meanwhile, the representation of different views from different instances should be distinctive to each other. To achieve this, one widely adopted method is contrastive learning~\cite{hadsell2006dimensionality}, which minimizes the similarity between views from the same instance while maximizes the similarity among views from different instances at the same time. To further improve contrastive learning, various methods~\cite{he2020momentum,chen2020simple,wu2018unsupervised,hjelm2018learning,oord2018representation,bachman2019learning,zhuang2019local,tian2019contrastive,henaff2019data} are proposed to explore different directions, such as the number of negative examples~\cite{wu2018unsupervised}, the quality of negative examples~\cite{he2020momentum,chen2020simple,hu2020adco}, data augmentation~\cite{tian2020makes,chen2020simple}.

On the one hand, it is worth noting that these methods usually rely on image augmentations that are carefully designated to maintain their instance identities so that the augmentation of an instance can be accurately retrieved from a pool of instances. The big impact of the data augmentation design has been clearly investigated in InfoMin~\cite{tian2020makes}, which also suggests the potential impacts of strong augmentations.  SwAV~\cite{caron2020unsupervised} and PIRL~\cite{misra2020self} also adopted stronger augmnetation compared to early work MoCo~\cite{he2020momentum} and SimCLR~\cite{chen2020simple}. However, no prior work has tried to apply random combinations of different augmentations together to have a much stronger augmentation like RandAugment~\cite{cubuk2020randaugment}. All those previous explorations still limit their focus on carefully artificially designed augmentations. 
On the other hand, novel patterns exposed by stronger augmentations can further boost the model's performance, which is clearly demonstrated in supervised and semi-supervised tasks~\cite{cubuk2020randaugment,cubuk2018autoaugment,wang2019enaet,sohn2020fixmatch}. Hence, we believe the patterns embedded in stronger augmentations could also contribute to self-supervised learning by improving the generalizability of learned representations and eventually close the gap with the fully supervised models. 
However, directly using stronger augmentations in contrastive learning could deteriorate the performance, because the induced distortions could severely change the image structures and thus the transformed images cannot keep the identity of the original instances. Thus, additional efforts are needed to explore the role of the stronger augmentations to further boost self-supervised learning.

Thus we propose the CLSA (\textbf{C}ontrastive \textbf{L}earning with \textbf{S}tronger \textbf{A}ugmentations)
 framework to address this challenge.  First, we introduced a much stronger augmentation named as "Stronger Augmentation" in this paper, which is a random combination of 14 types of augmentations: ShearX/Y, TranslateX/Y, Rotate, AutoContrast, Invert, Equalize, Solarize, Posterize, Contrast, Color, Brightness, Sharpness, whose details will be illustrated later. Instead of applying strongly augmented views to the contrastive loss, we propose to minimize the distribution divergence between the weakly and strongly augmented
images over a representation bank to supervise the retrieval of stronger queries. First, this design avoids an over-optimistic assumption that the strongly augmented view's embedding should be identical with that of the weakly augmented view. Meanwhile, leveraging weakly augmented counterparts' distributions enables our framework to explore the novel pattern carried by strongly augmented views. Most importantly, since it's independent to the contrastive loss, our framework can combine with any contrastive loss-based methods, such as MoCo~\cite{he2020momentum,chen2020improved}, SimCLR~\cite{chen2020simple}, BYOL~\cite{grill2020bootstrap} and so on. Our experiments have shown that our framework can greatly boost performance by introducing the distributional loss, which clearly suggests the novel distributional loss can explore the novel patterns exposed by the stronger augmentations and inherit the knowledge about the relative similarities to the negative samples. Furthermore, our experiments also verify that CLSA does not only improved the feature representation quality of weakly augmented views, but also further enhanced the representations of strongly augmented views at the same time. This property can greatly extend the self-supervised learning application on non-natural images, where the images are not carefully organized and the qualities of images cannot be guaranteed. 

The experiments on various datasets demonstrate that the proposed framework can significantly boost the performance by learning from stronger augmentations. On the ImageNet linear evaluation protocol, we reach a record 76.2\% top-1 accuracy with the standard ResNet-50 backbone, which is almost as high as 76.5\% top-1 accuracy of the fully supervised model. Meanwhile, it also achieves competitive performances on several downstream tasks. Among them is a top-1 accuracy of  93.6\% on VOC07 by the linear classifier with a pre-trained ResNet-50 compared to the previous record of 88.9\% top-1 accuracy. For the COCO object detection, the $AP_{S}$ for small object detection has been improved to 24.4\% from the previous best $AP_{S}$ of 20.8\%.  These results show that the CLSA can more effectively capture the semantic information than the previous self-supervised 
methods on downstream tasks by introducing the strongly augmented images. We also conduct an ablation study to show a naive application of stronger augmentations in contrastive learning would degrade the performances.

Our contribution can be summarized as follows.
\begin{itemize}
\item We are the first to explore the stronger augmentations to contribute to self-supervised learning.
\item We propose a distributional loss to transfer knowledge from weakly augmented views to strongly augmented views.
\item CLSA can easily integrate with concurrent contrastive loss-based methods and greatly boost their performance.
\item We carefully carry the ablation study to verify the impact of the distributional loss. 
\item CLSA framework can self-train network to improve representation for weakly augmented images and strongly augmented images simultaneously. 
\end{itemize}

\section{Related Work}
\label{sec:relawork}
\subsection{Self-Supervised Learning}
Self-supervised learning methods have been widely studied to close the gap with  supervised learning and alleviate the time and cost for labeling a large amount of data. These methods can be categorized into five different aspects.

\textbf{Instance Discrimination and Contrastive Learning} Each image is considered as an individual class in an instance discrimination setting \cite{hadsell2006dimensionality,wu2018unsupervised,chen2020simple,he2020momentum}. It can be further formulated as contrastive learning~\cite{hadsell2006dimensionality}, whose core idea is to pull the positive pairs together and push the negative pairs away in the embedding space. As studied in many prior works~\cite{hu2020adco,chen2020simple,chen2020improved,he2020momentum,wang2020unsupervised}, it is very crucial to construct high-quality positive and negative pairs to achieve higher performance. In particular, InstDisc\cite{wu2018unsupervised} built a memory bank that stores pre-computed representations as negative pairs to increase the size of the negative pair to improve the performance. Following this work, MoCo\cite{he2020momentum} used a momentum update mechanism to maintain a long queue of negative examples for contrastive learning. This momentum encoder design greatly improved the quality of negative pairs and led to a big performance increase compared to previous works.  SimCLR\cite{chen2020simple} further improved by directly utilizing negative samples in the current batch with a much bigger batch size. Meanwhile, it carefully constructed a rich family of data augmentations on cropped images, which significantly boosted classification accuracy. MoCo V2~\cite{chen2020improved} verified the improvement by introducing the same data augmentation and MLP layer design to MoCo framework. However, these methods failed to improve the performance by naively applying stronger augmentations to minimize the contrastive loss, which motivated the proposed work.

\textbf{Generative Methods} The generative methods typically adopt auto-encoders \cite{kingma2013auto}, and adversarial learning \cite{donahue2016adversarial} to train an unsupervised representation. Usually, they focused on the pixel-wise information of images to distinguish images from different classes. For instance, \cite{donahue2016adversarial} adopted BiGAN to capture the relationship between latent semantic representations and the input images.

\textbf{Clustering} Clustering~\cite{caron2018deep} can also be used to learn visual representations by assigning pseudo cluster labels to individual samples. DeepCluster~\cite{caron2018deep} generalized k-means by alternating between assigning pseudo-labels and updating networks. Recently, SWAV~\cite{caron2020unsupervised} is proposed to learn a cluster of prototypes by enforcing consistency between cluster assignments for different views, which has achieved state-of-the-art performance on ImageNet.

 Like contrastive learning, memory bank, large batches, and prototypes queue are also utilized in the clustering-based methods. Moreover, though clustering-based methods do not specifically use negative
pairs, the cluster centers can be viewed as negative prototypes. Hence, clustering-based approach can also be viewed as an extension of contrastive learning. 

\textbf{Consistency Representation Learning} Instead of directly using contrastive loss, researchers~\cite{grill2020bootstrap,chen2020exploring} found consistency representation learning between positive pairs can also enforce the network to learn robust representations. In BYOL~\cite{grill2020bootstrap}, researchers first found that we can self-train an encoder without using negative examples. It utilized the Siamese architecture, where the query branch has a predictor architecture in addition to the encoder and projector. The encoder can learn good representations by simply minimizing the cosine similarity between query embedding and key embedding. Following BYOL, Simsiam~\cite{chen2020exploring} further removed the momentum key encoder and used the stop-gradient strategy to avoid the collapsing problems. Furthermore, SCRL~\cite{roh2021spatially} further applied the consistency loss for ROI of the intersection region of two views to improve the encoder representation for downstream detection tasks. Concurrently, KL loss for consistency learning is widely used to help representation learning, such as CO2~\cite{wei2021co} and RELIC~\cite{mitrovic2020representation}, where regularization is added to enforce the consistency 
between the embeddings from different data augmentation. 

\textbf{Pretext Tasks} In addition to contrastive learning, there exist many alternative methods using different pretext tasks \cite{qi2019learning} to train networks with the carefully designed supervised signals from images. For example, \cite{noroozi2016unsupervised} proposed Jigsaw puzzles as supervised signal to train a convolutional
neural network. \cite{doersch2015unsupervised} used the relative positions of two randomly sampled patches as the supervised signal. \cite{zhang2019aet} used the transformations between two images as supervised signals to guide the representation learning. For more details about these works, please refer to the survey by \cite{jing2020self}.

\subsection{Augmentation in Representation Learning}

Data augmentation has played core rule in training deep neural networks. On the one hand, it helps the learned representation to be more robust under different data augmentation, which can help the model to learn transformation invariant representations. On the other hand, augmentation also introduced much more rich data for training.

In supervised learning, position and orientation adjustment was first introduced for MNIST~\cite{lecun1998gradient} dataset and achieved promising improvements~\cite{ciregan2012multi,simard2003best,wan2013regularization}. Later, for natural image dataset, such as CIFAR-10~\cite{krizhevsky2009learning}, ImageNet~\cite{deng2009imagenet}, random cropping, image mirroring and color shifting/whitening are introduced to train a better neural network~\cite{krizhevsky2012imagenet}. All those early works are 
manually designed, which requires time and expertise. When we wanted to combine data augmentation to have stronger augmentation, manual design is not feasible and optimal. To resolve this, researchers explore the combinations via two different approaches. 

First, Generative Adversarial Networks~(GAN)~\cite{antoniou2017data,mun2017generative,perez2017effectiveness} was used to directly generate more data with different transformations through generator. However, it's found that it~\cite{ravuri2019classification} is more beneficial to learn an optimal sequence of data augmentation through a pre-defined augmentation pool via a conditional GAN. Inspired this work, alternative approaches are proposed to carefully investigate how to automatically find a good data augmentation combinations. AutoAugment~\cite{cubuk2019autoaugment} first adopted reinforcement learning to learn the sequence of augmentation operations with application probability and magnitude. Following this work, Population Based Augmentation~(PBA)~\cite{ho2019population}, Fast AutoAugment~\cite{lim2019fast}, Faster AutoAugment~\cite{hataya2019faster} are proposed to accelerate the data augmentation policy search and improve it. RandAugment~\cite{cubuk2020randaugment} further found that uniform sampling of different data augmentations with uniformly sampled magnitude can build powerful data augmentation without extensive search. Those stronger augmentations by a combinations of different transformations greatly contribute to the supervised learning in both classification and detection.

In semi-supervised learning, MixMatch~\cite{berthelot2019mixmatch} introduced MixUp~\cite{zhang2017mixup} augmentation to help semi-supervised learning, where model is trained by mixed views and mixed labels with a convex combinations in MixUp. EnAET~\cite{wang2019enaet} utilized views with affine and projection transformation to further improve the semi-supervised learning. FixMatch~\cite{sohn2020fixmatch} found that highly distored images via stronger augmentation by RandAugment plays a key role to learn from a small amount of labeled data and large portion of unlabeled data.

In self-supervised learning, InstDisc~\cite{wu2018unsupervised} and MoCo~\cite{he2020momentum} added the color jittering operation in the data augmentation pipeline and had an obvious gain for contrastive learning.  SimCLR~\cite{chen2020simple} further added GaussianBlur to the data augmentation pipeline, where the improvement is further verified by MoCo v2~\cite{chen2020improved} and other following works. Based on those observations, InfoMin~\cite{tian2020makes} explored the effect of different combinations of different data augmentations in contrastive pre-training and has shown some combinations of data augmentation can have extra improvements. Meanwhile, SwAV~\cite{caron2020unsupervised} further introduced multi-crop with extra smaller 96*96 crops to help the model learn more robust feature representations. Also, BoWNet~\cite{gidaris2020learning} even included CutMix~\cite{yun2019cutmix} as stronger augmentation to contribute to the representation learning. All those previous work have explored how to design and build a more suitable data augmentation pipeline by introducing more and more transformations. However, they all require time, efforts and expert knowledge to manually design the augmentation and the designed data augmentation may only suitable for some datasets. To overcome that, we proposed stronger augmentation by a random combination of 14 types of augmentation with random probability of application and magnitude as well as the baseline augmentation in MoCo v2~\cite{chen2020improved}, where the details are illustrated in Sec.~\ref{sec:implementation}. First, the fully random combination of different augmentations build much stronger augmentation by repeating the sampling augmentation operation 5 times. Second, compared to previous approaches, our data augmentation is automatically randomly sampled without human intervention. In addition, as indicated in supervised learning~\cite{cubuk2020randaugment}  and semi-supervised learning~\cite{sohn2020fixmatch}, we also showed this much stronger augmentation helped a lot for the model to learn robust feature representations under distorted images.

\section{CLSA: Contrastive Learning with Stronger Augmentations}
In this section, we will first review the preliminary work on contrastive learning, and discuss their strength and limitations in Section.~\ref{sec:problem}.  Then in Section.~\ref{sec:ddm}, we will present a new distributional divergence loss between weakly and strongly augmented images to self-train the representations by leveraging the underlying visual semantic information from strongly augmented views.  After that, the implementation details 
 are explained in Section.~\ref{sec:implementation}.

\subsection{Contrastive Learning}
\label{sec:problem}
Contrastive learning \cite{hadsell2006dimensionality} is a popular self-supervised idea and made great success in recent years with the advance of computation and various image augmentations. Its goal is to find a parametric function $f_\theta$ that maps the input image $x\in \mathbb{R}^{D}$ to a feature representation $z=f_\theta(x) \in \mathbb{R}^{d}$, such that the feature representation $z$ in the feature space can reflect the semantic similarities in the input space. To this end, the contrastive loss is proposed to optimize the network $f_\theta$, which encourages $z$ and its positive pair $z'$ to be close in the feature space, and pushes away representations of all other negative pairs. After SimCLR~\cite{chen2020simple}, projector $g$ is imported to further map the representation as $z=g_\theta(f_\theta(x))$ for contrastive pre-training, while we still only use $f_\theta(x)$ for downstream tasks~(classification/detection). This design has been proved to the key to boost performance for contrastive learning~\cite{chen2020improved,chen2020simple}.

\begin{figure}[] 

\centering 
\includegraphics[width=0.48\textwidth]{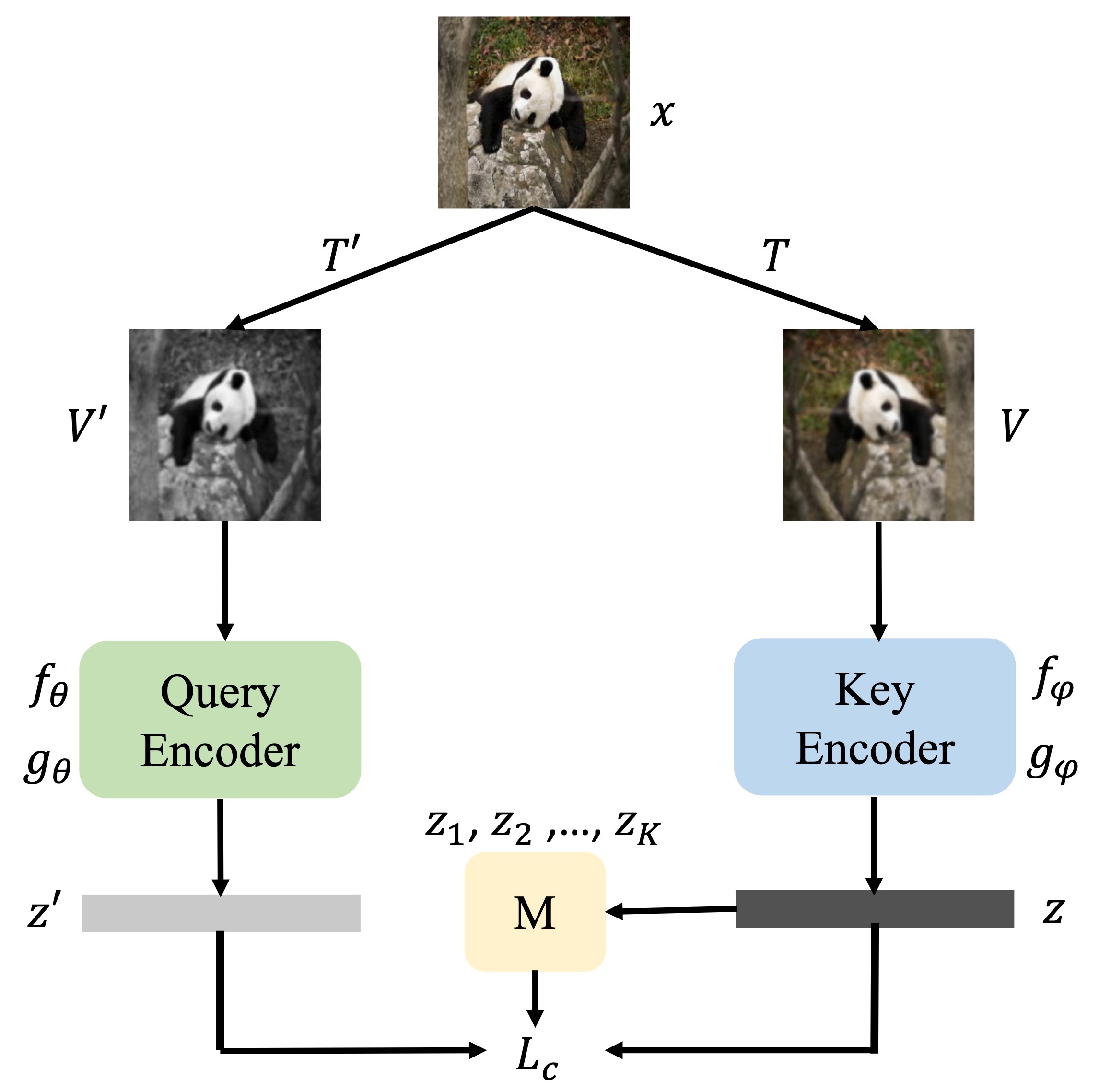} 

\caption{Contrastive instance learning framework} 
\label{fig:framework} 
\end{figure}

Fig.~\ref{fig:framework} illustrates the latest general framework of contrastive learning methods. In the supervised settings, the contrastive loss can be achieved by defining the same class images as positive pairs while remaining others as negative pairs. Similarly, the definition of positive pairs in contrastive loss is inspired by instance learning~\cite{hadsell2006dimensionality,wu2018unsupervised,he2020momentum,chen2020simple,chen2020improved}, where random augmented crops of the same image could be defined as positive pairs and crops from other images are all regarded as negatives. Hence, the contrastive loss in self-supervised learning is to maximize the agreement of representations of different views of the same instance while minimizing the agreement with other negative samples.

\begin{figure}[]
\centering 
\includegraphics[width=0.48\textwidth]{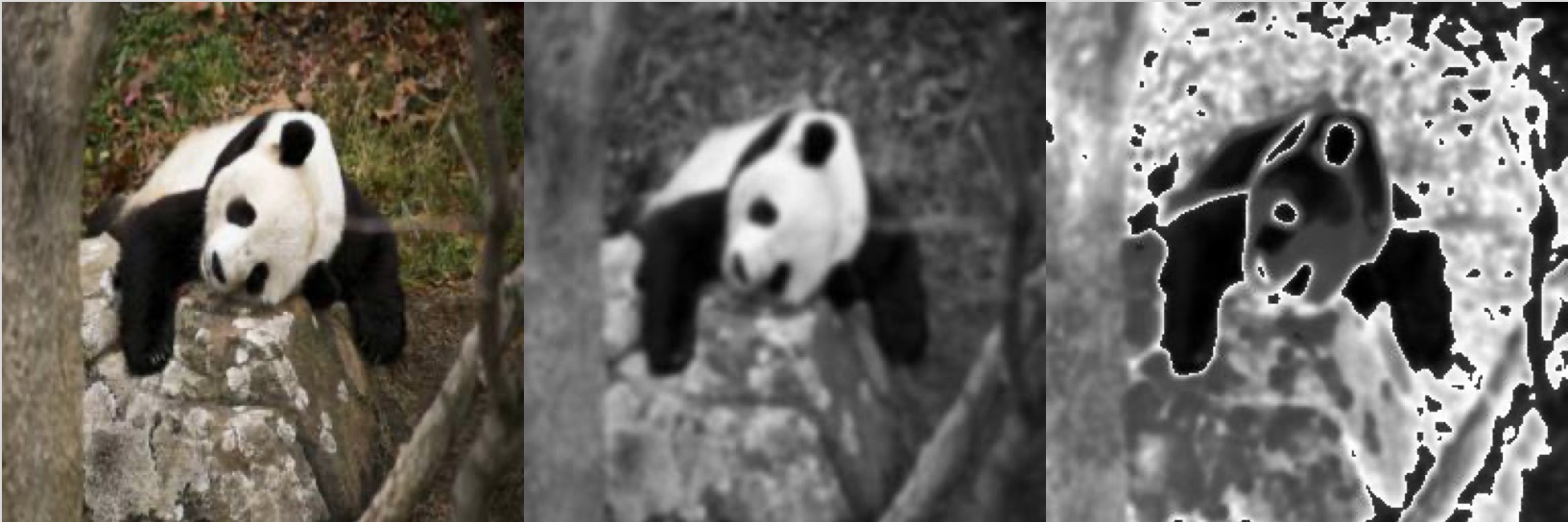} 
\caption{Comparison of the strongly weakly augmented images. The left is the original image, the middle is the weakly augmented image, and the right is the strongly augmented one with over-contrastive details. } 
\label{fig:strongexample} 
\end{figure}

Specifically, for each image $x$ in batch $B$, we apply two transformations $T'$ and $T$ to obtain two different views $V'$ and $V$ of the same instance $x$. Then they go through a query encoder $f_\theta$ and a key encoder $f_\phi$ respectively, followed with MLP projection layers ($g_\theta$/$g_\phi$), resulting in two embedded representations $z'$ and $z$ to calculate the constrastive loss in Eq.~(\ref{eq:contrastive}).
\begin{equation}
\label{eq:contrastive}
\centering
 \mathcal L_{C}=\mathbb{E}
 _{i\in B}[-
 \log \frac{Q(i,i+)}{Q(i,i+) + \sum_{k=1}^{K}Q(i,k)}]
\end{equation} 
with 
\begin{equation}
\label{eq:similarity}
\centering
\left\{  
\begin{array}{l} 
\mathcal Q(i,i+)=\exp(sim(z_{i}',z_{i})/\tau)\\
 \mathcal Q(i,k)=\exp(sim(z_{i}',z_{k})/\tau)\\
 sim(z_{i}', z_{k}) = \frac{z_{i}'^{T}z_{k}}{||z_{i}'||\cdot||z_{k}||}
\end{array}  
\right. 
\end{equation} 
where $\mathcal L_{C}$ is the contrastive loss, $Q(i,i+)$ is the exponential temperature smoothed similarities of positive pairs between $z_{i}'$ and $z_{i}$, while $Q(i,k)$ is the exponential temperature smoothed similarities of negative pairs between $z_{i}'$ and $z_{k}$, $\tau$ is the temperature parameter set to $0.2$, and $sim$ denotes the cosine similarity. Here $z_{k}$ is the feature embedding of other instances by $f_\phi$ and $g_\phi$. $K$ is the size of the FIFO queue $M$~(Memory Bank) saving the feature embeddings of other instances from previous batches.

In contrastive learning, success is highly dependent on two components: 

1) \textbf{The design of positive pairs}: For the positive pairs, the data augmentation is carefully designed. In SimCLR~\cite{chen2020simple}, it carefully designed color-jittering, Gaussian blurring transformations to further augment random cropped views. In InfoMin~\cite{tian2020makes} explored the effect of different data augmentations in contrastive pre-training and has shown some combinations of data augmentation can further improve compared to MoCo~\cite{chen2020improved} or SimCLR~\cite{chen2020simple}.

2) \textbf{The design of negative pairs}: For the negative pairs, researchers have explored a large number of approaches to improve the number and the quality of negative pairs. For example, InstDisc~\cite{wu2018unsupervised} first deployed memory bank~(negative pool) to keep track of feature embeddings of previous batches as negative pairs, which greatly improved the performance by a large pool of negatives. MoCo~\cite{he2020momentum,chen2020improved} further improved the quality of negatives by using momentum encoder as key encoder $g_\theta$. SimCLR~\cite{chen2020simple} kept a balance of quality and number of negatives by utilizing large batch online training and using other instances in the same batch serving as negatives for optimizing contrastive loss. 
Recently, AdCo~\cite{hu2020adco} even utilized an adversarial memory bank to serve as negative pairs, where the memory bank can be trained to generate negative features by end-to-end training. In short, the potential of negatives have been fully explored in various perspectives and greatly improved representation learning.

Nevertheless, the construction of positive pairs has not received enough attention and studies yet. As aforementioned, the two views of the same instance in contrastive learning are not transformed aggressively so that they can still be viewed as the same instance. InfoMin~\cite{tian2020makes} has proved  model can learn more information by different combinations of weak augmentations.
That inspired us that augmentations for positive pairs can leverage possible semantic information for the encoder to learn. Furthermore, we do not want to rely on the carefully designed augmentation for training. Instead, we aim to explore a general and robust approach to learn from infinite data augmentations.

However, directly adopting stronger transformations (e.g., with larger rotation angles, more aggressive colorjittering and cutout) in contrastive learning fails to further improve the performance or even deteriorate it for downstream tasks, which is not surprising.  Stronger transformations could distort image structures and their perceptual patterns in the learned representation so that strongly augmented counterpart views cannot be viewed as the same instance for training the underlying network. In InfoMin~\cite{tian2020makes}, they also only explored the combinations of weak augmentations instead of stronger augmentations that may include more information to help model to learn robust features.Since different combinations of weak augmentations may provide different information for obtaining distinctive feature representations, it's highly possible that some helpful information can only be learned via stronger augmentation. In supervised learning~\cite{cubuk2020randaugment,cubuk2019autoaugment,hataya2020faster} and semi-supervised learning~\cite{wang2019enaet}, different stronger data augmentations have been widely studied and greatly boost the performance with the novel pattern exposed by strongly augmented images. The findings in RandAugment~\cite{cubuk2020randaugment} have verified that strongly augmented views can provide more clues even without an explicit augmentation policy. Hence, we believe learning the representations from these novel patterns will pave the last mile to close the gap with the fully supervised representations. That further inspired us to explore novel ways to utilize stronger transformations in self-supervised learning while avoiding deteriorated performances by naively using them in a contrastive model~\cite{chen2020improved}.

\begin{figure}[]
\centering 
\includegraphics[width=0.48\textwidth]{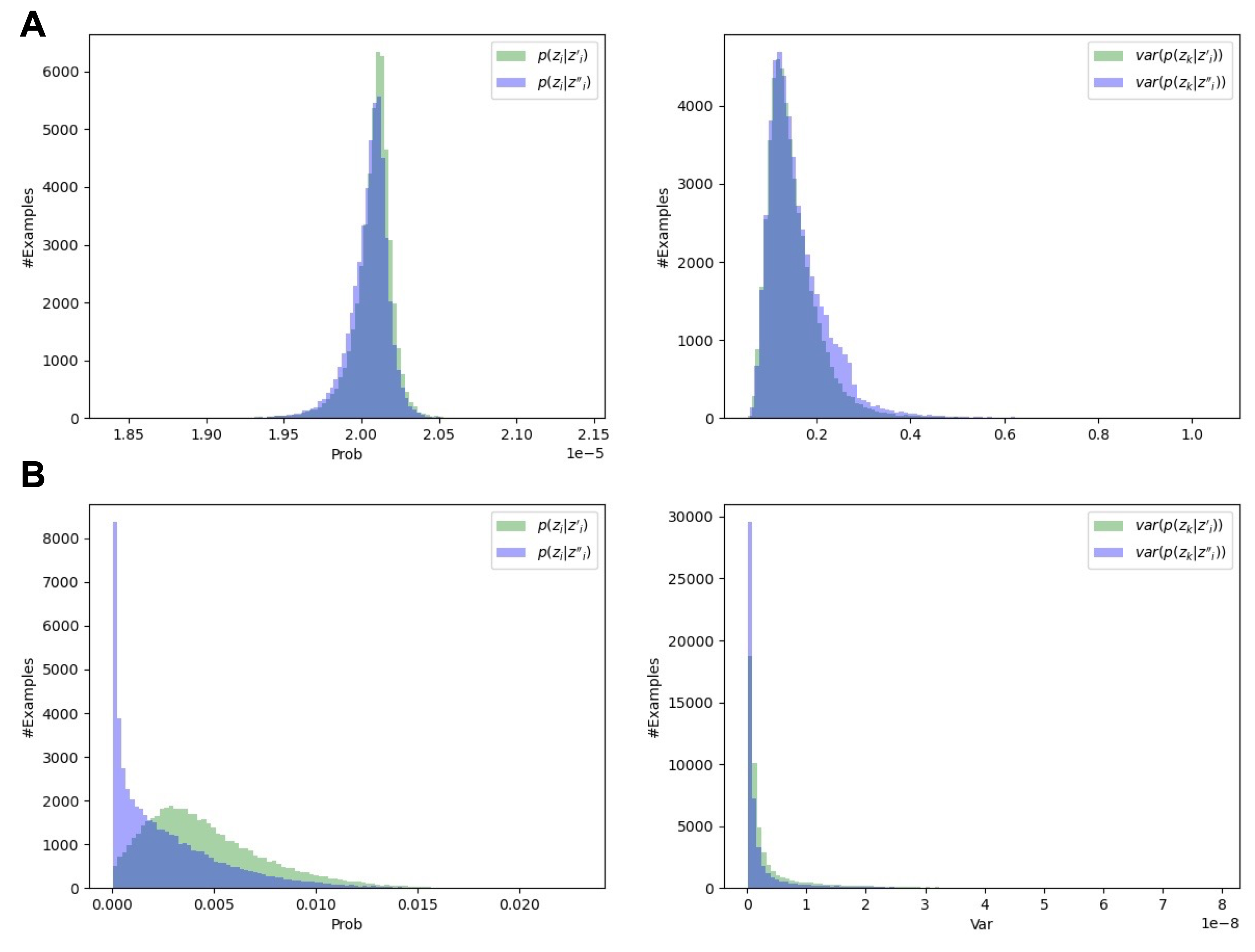} 
\caption{The comparison of the distribution of positive pair's probabilities and variance of negative pairs' probabilities given weakly augmented query and strongly augmented query. \textbf{A.} The distribution with a randomly initialized network.  \textbf{B.} The distribution with a pre-trained network by contrastive methods.} 
\label{fig:clsamotivation} 
\end{figure}

By exploring previous approaches and our extensive experiments, we have found it is not a straightforward task to learn the patterns embedded in the stronger augmentations.  As shown in Fig.~\ref{fig:strongexample}, a strongly augmented image may look perceptually different from the original counterpart. Consequently, the representation of a strongly augmented image can be far apart from that of the weakly augmented one. Thus, naively using strongly augmented images in contrastive learning can be over-optimistic since the induced distortions could dramatically change their image structures.

To this end, in Section \ref{sec:ddm}, we instead proposed the Distributional Divergence Minimization (DDM) between weakly and strongly augmented images over a representation bank to avoid overfitting the representation of a strongly augmented image with that of the corresponding positive target.

\subsection{Distributional Divergence Minimization between Weakly and Strongly Augmented Images}
\label{sec:ddm}

\begin{figure*}[] 
\centering 
\includegraphics[width=\textwidth]{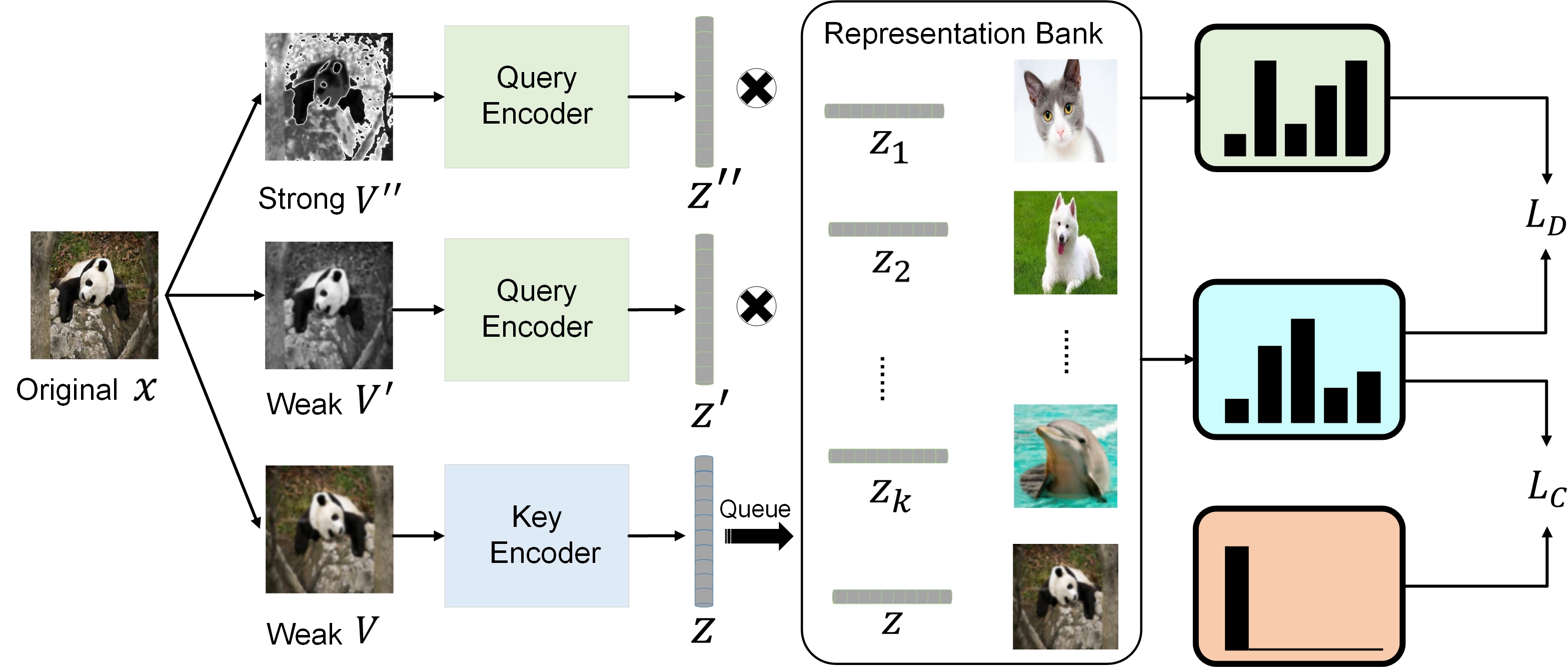} 
\caption{Diagram of distributional divergence minimization. Here the representation bank consists of $K$ stored features $z_{k}$ of previous batches and online features $z_{i}$ from the key encoder. They will be used to calculate the conditional probability of current weakly and strongly augmented query images.} 
\label{fig:ddm} 
\end{figure*}

Due to the limitation mentioned above, learning from the retrieval of a strongly augmented query is infeasible to self-train deep networks. Nevertheless, the distribution of relative similarities can help us understand contrastive learning in a different direction, thus inspired us to propose the distributional divergence minimization (DDM) for learning from stronger augmentations.

As shown in Fig.~\ref{fig:framework}, for an image $x$, we can first apply a transformation $T'$ to generate a query view $V'$ and then obtain its corresponding embedding $z'=g_\theta(f_\theta(V'))$. Meanwhile, we collect another key view $V$ generated by $T$ and the key embedding $z=g_\phi(f_\phi(V))$. 
Given a memory bank $M$ of $K$ negative samples $\{z_k|k=1,\cdots,K\}$ accumulated from key embedding $z$ of the past iterations, we can obtain a conditional distribution 
\begin{equation}
\label{eq:prob_relative}
\centering
p(z_{j}|z'_{i})=\frac{\exp(sim(z'_{i},z_{j})/\tau)}{\exp(sim(z'_{i},z_{i})/\tau)+\sum_{k=0}^{K}\exp(sim(z_{i}',z_{k})/\tau)}
\end{equation}
which encodes the likelihood of the query $z'_i$ being assigned to the embedding $z_{j}$. Hence $p(z_{j}|z'_{i})$ represents the positive likelihood when $j=i$, and it also represents its relative negative likelihood with $z_{k}$ when $j=k$, where $z_{k}$ is a negative embedding from the memory bank $M$.

Then, the contrastive loss in Eq.~(\ref{eq:contrastive}) can be rewritten in another form as

\begin{equation}
\label{eq:contrastive2}
\centering
 \mathcal L_{C}=\mathbb{E}
 _{i\in B}[-q(z_i|z_{i}')\log p(z_i|z_{i}')-\sum_{k=1}^{K}q(z_k|z_{i}')\log p(z_k|z_{i}')]
\end{equation} 
where $q(z_i|z_{i}')$ is the ideal distribution of the likelihood, $p(z_i|z_{i}')$ is the distribution learned by network.Here the ideal distribution denotes the likelihood the representation of a query assigned to different keys according to semantic similarity. However, we can't have a precise measurement to identify it precisely. 

However, we cannot obtain the ideal distribution of the likelihood from the semantic view. It is hard to measure the optimal likelihood between the query image and the key images~(positive/negative). To avoid this unknown distribution exploration, contrastive loss regards $q$ as a one-hot distribution, where positive pair has $q(z_i|z_{i}')=1$ and negatives satisfy $q(z_k|z_{i}')=0~(k\in[1,K])$. That means contrastive loss only maximizes the agreement of representations of different views of the same instance while minimizing the agreement with other negative samples. All other underlying complicated relationships between query images and key images are completely ignored. The advantage is that contrastive loss can greatly accelerate the convergence of representation learning and greatly improve the representation feature for classification and detection tasks. However, the information between the query image and negative images is not fully explored and may include useful clues to boost representation learning further. 

Similar to the representation for weakly augmented views, a straightforward solution for exploring the patterns from the stronger augmentation is directly using the strongly augmented image as query and using the weakly augmented image as key in the contrastive loss. However, this too optimistic design preassumes that the strongly augmented view's representation should be close to that of its weakly augmented pair and far from that of the weakly augmented views of other instances. The one-hot distribution fails to mimic or even approximate the optimal likelihood distribution and thus cannot help the representation learning anymore. Hence, another alternative distribution $q$ should be proposed to address these limitations of one-hot distribution. 

 Though it is almost impossible to obtain the actual likelihood distribution to self-train network in a perfect way, fortunately, we found that the distribution of relative similarities of a weakly augmented image from the same instance over the representation bank can provide useful clues for stronger augmentation learning. In Fig.~\ref{fig:clsamotivation}, we compared the distribution of positive pair's probabilities $p(z_i|z'_i)$~($p(z_i|z''_i)$) and the distribution of variance of negative pairs' probabilities $p(z_k|z'_i)$~($p(z_k|z''_i)$) for $k\in[1,K]$ for weakly~(strongly) augmented query. In Fig.~\ref{fig:clsamotivation}A, the initial similarity distribution of strongly augmented query is identical to that of weakly augmented query. That indicates there exist no difference of embedding of strongly views or weakly views with a pre-trained network. However, after training with one of the most representative contrastive-based approaches MoCo~\cite{he2020momentum,chen2020improved}, the difference of similarity distribution between the strongly augmented view and weakly augmented view becomes obvious as shown in Fig.~\ref{fig:clsamotivation}B. In other words, the distribution difference suggests that concurrent contrastive methods failed to learn representations that are robust to stronger image distortion. That inspired us to propose an approach that can learn the representations that are stable under stronger augmentations. Meanwhile, we can't directly pull the representation of strongly augmented view to that of the weakly augmented view, since we found that will destroy the representation learning through our experiments. As an alternative, we adopted a relaxed protocol to utilize the distribution of relative similarities between query and key to pre-train the model.  
  This property inspired us that the distribution of relative similarities of weakly augmented query can be used to supervise that of the strongly augmented query. 
 
 Our ablative studies found that the representations of strongly augmented views by contrastive methods are much less distinctive than that of weakly augmented views. Thus, we believe the distribution supervision of weakly augmented views can simultaneously 
stabilize the representation of strongly augmented views.

In a nutshell, it does not only avoid directly placing the representation of a strongly augmented image too closely to that of the positive target~(like one-hot distribution did), but also allows it to explore the novel patterns of variations exposed by the strong augmentation.

Formally, as shown in Fig.~\ref{fig:ddm}, for an original image $x_i$, we applied another stronger augmentation $S$ to obtain its strongly augmented view $V''_i$, and its embedding $z''_i$. Meanwhile,  we can also obtain its corresponding query embedding $z'_{i}$ and key embedding $z_{i}$ for the instance $x$ by weak augmentations $T'$ and $T$.  Similar to Eq.~(\ref{eq:prob_relative}), we obtain a conditional distribution for $z''_i$ based on its positive target and negative pairs:
\begin{equation}
\label{eq:transformer}
\centering
p(z_{j}|z''_{i})=\frac{\exp(sim(z''_{i},z_{j})/\tau)}{\exp(sim(z''_{i},z_{i})/\tau)+\sum_{k=0}^{K}\exp(sim(z_{i}'',z_{k})/\tau)}
\end{equation}
where $p(z_{j}|z''_{i})$ indicates the positive likelihood when $j=i$, and it can also reflect the negative likelihood when $j=k$, where $z_{k}$ is a negative embedding from memory bank $M$.

Then, we propose to minimize the following distributional divergence between the weak and the strong queries such that
\begin{equation}
\label{eq:DDM}
\centering
 \mathcal L_{D}=\mathbb{E}
 _{i\in B}[-p(z_i|z_{i}')\log p(z_i|z_{i}'')-\sum_{k=1}^{K}p(z_k|z_{i}')\log p(z_k|z_{i}'')]
\end{equation}
By minimizing this divergence, we assume the learned representation $z''_i$ of the strongly augmented query should inherit the representation $z'_i$ of the weakly augmented one regarding not only its belief of the query being assigned to the corresponding positive target $z_i$, but also its  relations with the negative samples $z_k$ in the representation bank through the conditional distribution $p(z_k|z'_i)$. We believe the distribution $p(z_k|z_{i}')$ is a much better mimic of $q(z_k|z_{i}'')$ instead of the one-hot distribution, which is also validated in our ablation study. Here we apply stop-gradient for $p(\cdot|z_{i}')$ and the gradient is only propagated through $p(\cdot|z_{i}'')$ branch.

\begin{table}[]
\centering
\caption{Various augmentations we applied in experiments to strongly augment training images.}
\label{tab:augmentation}
\resizebox{0.49\textwidth}{!}{
\begin{tabular}{lcccccc}
\toprule[1pt]
Operation & ShearX(Y) & TranslateX(Y) & Rotate \\ 
Mag Range & [-0.3,0.3] & [-0.3,0.3] & [-30,30] \\\hline
Operation & AutoContrast &Invert &Equalize\\ 
Mag Range &  0 or 1 & 0 or 1 & 0 or 1\\\hline
Operation & Solarize & Posterize&Contrast \\
Mag Range & [0,256] & [4,8] &[0.05,0.95]\\ \hline
Operation & Color&Brightness &Sharpeness\\
Mag Range &[0.05,0.95]&[0.05,0.95]&[0.05,0.95]\\
\bottomrule[1pt]
\end{tabular}}
\end{table}

This will prevent direct overfitting of the strong query representation to the positive target as well as improve the generalization of the learned representation with additional clues from the other examples in the representation pool. In a more general sense, this extends the idea of knowledge distillation~\cite{hinton2015distilling}.  However, we did not use the predicted labels by a teacher model to supervise a student model's training as in the knowledge distillation. Instead, we used the distribution of the likelihoods of a weak query to supervise the retrieval of a strong query from a pool of representations.

Since the distributional divergence optimization can be viewed as a knowledge distillation process, that also indicates that the success is highly dependent on the distribution of relative similarity of weakly augmented query $z'$. Hence, it is natural to utilize one of the current contrastive methods to learn $z'$ through the contrastive loss. That design also enables our DDM loss to be fully independent with concurrent contrastive methods and can complement any of them to further boost them. As illustrated in Fig.~\ref{fig:ddm}, the overall loss to optimize the encoder can be formulated as
\begin{equation}
\label{eq:overallloss}
\centering
 \mathcal L= \mathcal L_{C}+\beta*\mathcal L_{D}
\end{equation}
where $\beta$ is the coefficient to balance the contrastive loss $\mathcal L_{C}$ and distributional divergence minimization loss $L_{D}$. Though other values may achieve better results for different contrastive baselines, we use $\beta=1$ to make CLSA more general .

As shown in Fig.~\ref{fig:ddm}, the pattern from strongly augmented images and weakly augmented images are learned simultaneously. That is a very natural and effective design since contrastive loss can also be viewed as a specific form of distributional divergence minimization as we have shown before, where the supervised distribution is a one-hot distribution. Concurrently, DDM loss is also used to help contrastive loss for consistency learning~\cite{wei2021co}
between the embeddings of two weakly augmented views and
invariant predictions under different data augmentation~\cite{mitrovic2020representation}. All
those works with DDM loss can help the model to learn better representations. Here, we can further found more improvements with the distillation between weakly and strongly augmented views.
In short, the overall optimization thus can also be interpreted as two simultaneous distributional divergence minimization. The extraordinary performance also supported this simple yet effective design.

\begin{table}
\centering
\caption{Top-1 accuracy under the linear evaluation on ImageNet with the ResNet-50 backbone with 200 epochs training. }
\label{tab:imagenet}

\begin{tabular}{ll}
\toprule[1pt]
Method & Top 1 \\ \toprule[1pt]
InstDisc~\cite{wu2018unsupervised} & 54.0 \\
LocalAgg~\cite{zhuang2019local} & 58.8 \\
MoCo~\cite{he2020momentum} & 60.8 \\
SimCLR~\cite{chen2020simple} & 61.9 \\
CPC v2~\cite{henaff2019data} & 63.8 \\
PCL~\cite{li2020prototypical} & 65.9 \\
MoCo v2~\cite{chen2020improved} & 67.5 \\
InfoMin Aug~\cite{tian2020makes} & 70.1\\
SWAV~\cite{caron2020unsupervised} (Multi-Crop)& 72.7 \\ \toprule[1pt]
CLSA & 69.4 \\
CLSA* (Multi-Crop) & \textbf{73.3} \\ 
Supervised & 76.5\\
\bottomrule[1pt]
\end{tabular}
\end{table}

\subsection{Implementation Details}
\label{sec:implementation}

In this section, we will discuss the details about the applied strong and weak augmentations for CLSA.

\textbf{Stronger Augmentations $\bm S$} As explored in the previous works (\cite{cubuk2018autoaugment,wang2019enaet,qi2019learning}), strong augmentations usually have two types: geometric and non-geometric augmentations. Specifically, we considered $14$ types of augmentations: ShearX/Y,
TranslateX/Y, Rotate, AutoContrast, Invert, Equalize, Solarize, Posterize, Contrast, Color, Brightness, Sharpness.  The magnitude of each augmentation is significant enough to produce as strong augmentations as possible. More details of different transformations are shown in Table~\ref{tab:augmentation}. For example, the shear operation is drawn from a range of [-0.3,0.3], resulting in aggressively transformed images that can be hard to retrieve given a counterpart target. In particular, to transform an image, we randomly select an augmentation from the above $14$ types of transformations, and apply it to the image with a probability of $0.5$.  This process is repeated five times and that will strongly augment an image as the example shown in the right panel of Fig.~\ref{fig:strongexample}. Compared to the weakly augmented image in the middle panel, it is clear the image structures of the strongly augmented views are completely changed.

\textbf{Weaker Augmentations} $\bm T$ Weak augmentations are drawn by following most of existing contrastive learning methods in literature (\cite{chen2020simple,chen2020improved,caron2020unsupervised,he2020momentum}): an image is first cropped from an input image and resized to 224$\times$224 pixels.
Then random color jittering, Gaussian Blur, grayscale conversion, horizontal flip, channel-wise color normalization are sequentially applied to generate weakly augmented images with an example shown in the middle of Fig.~\ref{fig:strongexample}. 

\textbf{Technical Details} Similar to the previous works (\cite{he2020momentum,chen2020simple,caron2020unsupervised}), we used the ResNet-50 (\cite{he2016deep}) as our encoder backbones $f_{\theta}$ and $f_{\phi}$ and a 2-layer MLP  (2048-d hidden layer with the ReLU, output FC without ReLU) as the projection head $g_{\theta}$ and $g_{\phi}$. The cosine similarity is used in the contrastive loss and DDM loss. The temperature $\tau$ is set to $0.2$. Following MoCo~\cite{he2020momentum}, a momentum smoothing factor $\alpha$ of 0.999 is used to update key encoder $f_\phi=\alpha*f_\phi+(1-\alpha)*f_\theta$ and key MLP $g_\phi=\alpha*g_\phi+(1-\alpha)*g_\theta$.  The loss balancing coefficient $\beta$ is set as $1.0$. 
We set the size $K$ of the queue $M$ to $65536$ to store the negative examples used to compute the conditional distribution of weakly and strongly augmented queries and minimize their divergence. We used the same temperature for DDM loss and contrastive loss to simplify the formulation. We believe the performance can be further improved by adjusting different temperatures for $\mathcal L_{C}$ and $\mathcal L_{D}$.
\begin{table}
\centering
\caption{Top-1 accuracy under the linear evaluation on ImageNet with the ResNet-50 backbone with various numbers of epochs.}
\label{tab:imagenet2}
\begin{tabular}{lll}
\toprule[1pt]
Method & Epoch& Top 1 \\ \toprule[1pt]
BigBiGAN~\cite{donahue2019large} & 2048 & 56.6 \\
SeLa~\cite{asano2019self} &400& 61.5 \\
PIRL~\cite{misra2020self}&800 & 63.6 \\
CMC~\cite{tian2019contrastive}& - & 66.2\\
SimCLR~\cite{chen2020simple}&800 & 70.0 \\
MoCo v2~\cite{chen2020improved}&800 & 71.1 \\
InfoMin Aug~\cite{tian2020makes} &800& 73.0\\
BYOL~\cite{grill2020bootstrap}&1000&74.3\\
SWAV~\cite{caron2020unsupervised}  (Multi-Crop) &800& 75.3 \\
 \toprule[1pt]
CLSA&800 & 72.2\\
CLSA*~(Multi-Crop) &800 & \textbf{76.2} \\
Supervised & -&76.5\\
\bottomrule[1pt]
\end{tabular}
\end{table}

\begin{table*}[!htb]
\centering
\caption{Transfer learning results on various downstream tasks.}
\label{tab:downstream}
\begin{tabular}{lccccc}
\toprule[1pt]
 & Classification & \multicolumn{3}{c}{Object Detection} \\ 
 \cline{2-5}
 & VOC07 & VOC07+12 & \multicolumn{2}{c}{COCO} \\
Measurement & Accuracy & $AP_{50}$ & AP & $AP_{S}$ \\ \toprule[1pt]
RotNet~\cite{gidaris2018unsupervised} & 64.6 & - & - & - \\
NPID++~\cite{wu2018unsupervised} & 76.6 & 79.1 & - & - \\
MoCo~\cite{he2020momentum} & 79.8 & 81.5 & - & - \\
PIRL~\cite{misra2020self} & 81.1 & 80.7 & - & - \\
PCL~\cite{li2020prototypical} & 84.0 & - & - & - \\
BoWNet~\cite{gidaris2020learning} & 79.3 & 81.3 & - & - \\
SimCLR~\cite{chen2020simple} & 86.4 & - & - & - \\
MoCov2~\cite{chen2020improved} & 87.1 & 82.5 & 42.0 & 20.8 \\
SWAV~\cite{caron2020unsupervised} & 88.9 & 82.6 & 42.1 & 19.7 \\ \toprule[1pt]
CLSA & \textbf{93.6} & \textbf{83.2} & \textbf{42.3} & \textbf{24.4} \\
Supervised & 87.5 & 81.3 & 40.8 & 20.1 \\ \bottomrule[1pt]
\end{tabular}
\end{table*}

\section{Experiments}
\label{label:experiments}

\subsection{Training Details}
For the unsupervised pretraining on ImageNet with the CLSA, we used the SGD optimizer~(\cite{bottou2010large}) with an initial learning rate of 0.03, a weight decay of 0.0001 and a momentum of 0.9. We used cosine scheduler~(\cite{loshchilov2016sgdr}) to gradually decay the learning rate to 0. Usually, the batch size is set to 256. When multiple GPU cluster servers are used, the batch size will be multiplied by the same number of servers by convention. To avoid collapsing, the shuffle BN~\cite{he2020momentum} is also adopted. The learning rate then will be adjusted based on $lr=base\_lr*batch\_size/256$, where $base\_lr$ is 0.03.
Typically, the experiment with a single strong augmentation for each training image takes roughly 70 hours to finish on 8 V100 GPUs. 

For the fine-tuning on ImageNet, we trained a linear classifier on top of  the frozen feature vector (2048-D) upon the pre-trained ResNet-50 with CLSA. This linear layer is trained for 100 epochs, with a learning rate of 10 without weight decay. We used the cosine learning rate decay to 0.01 and a batch size of 256. 

For the transfer learning on the VOC dataset, we trained a linear classifier upon the pre-trained Resnet-50 in a similar way for ImageNet -- we trained 100 epochs with the SGD optimizer and a learning rate of 0.05, a momentum of 0.9 and no weight decay. The batch size is 256 without a learning rate scheduler.

Finally, for object detection, we adopted the same protocol in (\cite{he2020momentum}) to fine-tune the pre-trained Resnet-50 backbone based on detectron2~(\cite{wu2019detectron2}) for the sake of a fair and straight comparison with the other methods.  

\subsection{Linear Classification on ImageNet}
\label{sec:linear}

For the linear evaluation on ImageNet, we trained the CLSA in two settings. In the first setting named CLSA, we used a single stronger augmentation (see Table~\ref{tab:augmentation}) that crops each training image to a smaller size of $96\times96$, which does not incur too much computing overhead in processing these smaller augmented images. In the second setting named CLSA*, we adopted five different stronger augmentations that crop each image into various sizes: $224\times224,192\times192,160\times160,128\times128,$ and $96\times96$. The DDM loss in Eq.~(\ref{eq:DDM}) is the sum over these multiple stronger augmentations. A similar multi-crop strategy has been adopted in contrastive learning literature before. For example, the SWAV~\cite{caron2020unsupervised} reached  state-of-the-art top-1 accuracy by applying such multi-crop augmentations. To ensure a fair comparison with the SWAV, we chose five stronger augmentations such that the self-training with CLSA* consumed similar computing time (i.e., 166 hours with a cluster of 8 V100 GPUs for 200 epochs of pre-training with a batch size of 256).

\begin{table}[]
\centering
\caption{Ablation study of the CLSA on ImageNet with 200 epochs of pre-training.}
\label{tab:ablation}
\begin{tabular}{ll}
\toprule[1pt]
Model & Top-1 \\ \toprule[1pt]
MoCo V2 & 67.5 \\
MoCo V2 with Strong query & 67.7 \\
MoCo V2 with Strong query \& Strong key & 67.0 \\
CLSA with contrastive loss & 68.0 \\ \hline
CLSA & \textbf{69.4} \\ \toprule[1pt]
\end{tabular}
\end{table}

As shown in Table~\ref{tab:imagenet} and Table~\ref{tab:imagenet2}, we compared the performance with the other unsupervised methods. All the experiments are based on a pre-trained ResNet-50 backbone that is fine-tuned with a linear classifier. Table~\ref{tab:imagenet} showed the performance of different methods pre-trained over 200 epochs, and table~\ref{tab:imagenet2} reported models pre-trained over more epochs.

First, under the same contrastive protocol, the CLSA has a higher $69.4\%$ top-1 accuracy than both MoCo v2 (67.5\%) and SimCLR (61.9\%) with 200 epochs training. 
With multiple stronger augmentations, CLSA* outperforms the state-of-the-art SWAV model using multi-crops of training images over 200 epochs (73.3\% vs. 72.7\%). Moreover, as shown in ~\ref{tab:imagenet2}, the CLSA outperforms MoCo v2 and SimCLR with the same training epochs. It is also noteworthy that the CLSA* achieves almost the same top-1 accuracy as that of the fully supervised network (76.2\% vs. 76.5\%). With the advance of more contrastive learning approaches, we believe CLSA can work with them to further improve their performance to beat supervised learning performance with more robust feature representations.

\subsection{Transfer Learning Results on Downstream Tasks}
\label{sec:transfer}

We test the generalizability of the  ResNet-50 pre-trained on ImageNet to several downstream tasks. Specifically, we focused on two tasks: cross-dataset image classification and object detection. The pre-trained ResNet-50 was frozen, and we fine-tuned the linear classifier on the VOC07trainval and tested it on the VOC07test. For object detection, we evaluated the pre-trained network on two datasets using the detectron2~\cite{wu2019detectron2} used in the previous methods~\cite{he2020momentum,chen2020simple}. On the VOC dataset, we trained the detection head with VOC07+12 trainval dataset and tested it on VOC07 test dataset. We fine-tuned the network on the train2017 set with 118k images on the COCO dataset and evaluated it on the val2017. For the sake of a fair comparison, the object detection tasks are completed by detectron2 based on the pre-trained ResNet-50.

As shown in Table~\ref{tab:downstream}, the performances on both tasks are much better than the supervised model trained on ImageNet. This suggests that the proposed method has better generalization ability in downstream tasks. The pre-trained network on ImageNet by the CLSA outperformed the compared models after being fine-tuned on different datasets. Among them is a top-1 accuracy of 93.6\% on the VOC07 with the linear classifier on the pre-trained ResNet-50 in comparison with the previous record of 88.9\% top-1 accuracy by the SWAV. On the COCO  dataset, the $AP_{S}$ for small object detection has been significantly
improved to 24.4\% from the previously best $AP_{S}$ of 20.8\%. As well known, it is much more challenging to detect small objects on the COCO dataset. Thus, the better performance of the CLSA could be attributed to the ability to involve the stronger augmentations that result in many small objects to pre-train the network.

\begin{table}[]

\centering
\caption{Training Time Comparison of CLSA}
\label{tab:ablation2}
\centering
\begin{threeparttable}
\begin{tabular}{llll}
\toprule[1pt]
Model & Time & Epoch& Top-1\\ \toprule[1pt]
MoCo V2 & 53h & 200& 67.5 \\
CLSA & 35h & 100 &67.2\\ 
CLSA & 52.5h & 150 &68.3 \\
CLSA & 70h & 200 & 69.4 \\ \toprule[1pt]
\end{tabular}
\begin{tablenotes}
\item[1] $h$ denotes running hours on a machine with 8 V100 GPUs.
\end{tablenotes}
\end{threeparttable}
\end{table}

\subsection{Ablation Study}
\label{sec:ablation}

In this section, we analyzed the contribution of DDM loss by comparing it with other designs. Also, we compared CLSA with other baselines using the same running time for a more fair comparison. Moreover, we compared CLSA performance under different strong augmentations. In addition, we carefully addressed our feature representation improvement for both weakly augmented views and strongly augmented views. Last but not least, we further studied the distribution of probabilities of positive and negative pairs with pre-trained network by CLSA. 

\textbf{DDM Loss} In the ablation study shown in Table~\ref{tab:ablation}, we studied the role of the proposed DDM loss in the CLSA. First, we naively used the stronger augmentation applied in the CLSA-Single as the query and/or the key in the MoCo V2. Both results (Strong query and Strong query \& Strong key) showed the performance could not be improved or even degraded.  Second, we replaced the DDM loss in the CLSA-Single with the contrastive loss, which can be viewed as replacing $p(z_{i}|z'_{i})$ and $p(z_{k}|z'_{i})$ with a one-hot distribution, and we found it can only achieve a top-1 accuracy of 68.0\% compared to that of 69.4\% with the DDM loss. That clearly indicates $p(z_{k}|z'_{i})$ is a more suitable approximation of $q(z_{k}|z''_{i})$ compared to one-hot distribution. Both studies showed that the proposed CLSA and its DDM loss help us learn from stronger augmentations by avoiding the performance degeneration that augmented images' distortions would incur.

 \begin{figure}[] 
\centering 
\includegraphics[width=0.5\textwidth]{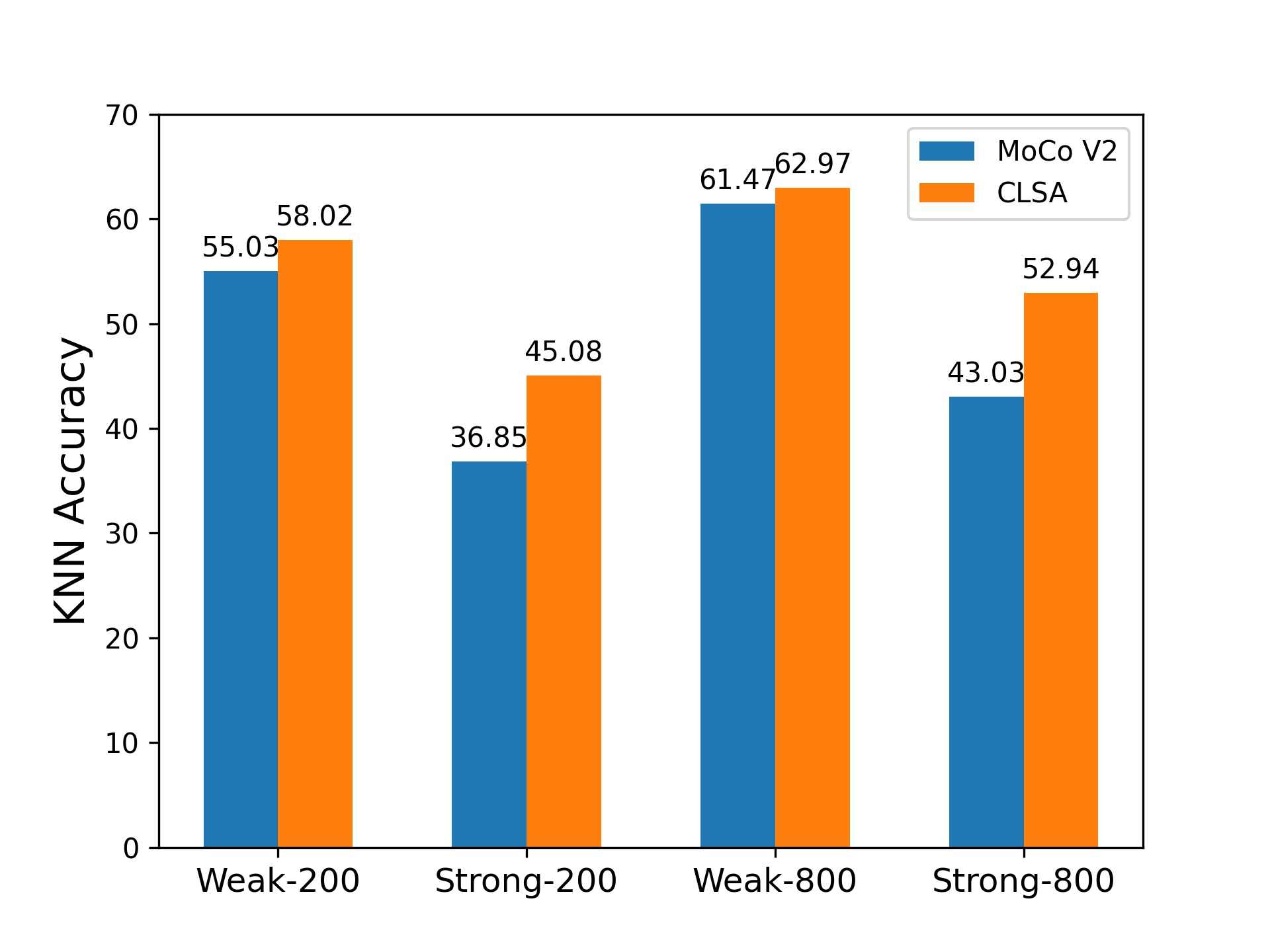} 
\caption{Comparison of KNN accuracy of MoCo V2~\cite{chen2020improved} and CLSA. Both results are compared based on pretrained model with 200/800 epochs with single crop. The comparison used the representation of weakly/strongly augmented images, respectively. The neighbor $K$ for KNN here is set to 20.} 
\label{fig:knn} 
\end{figure}

\textbf{Running Time} To address the concern of extra training time consuming of CLSA compared to MoCo V2~\cite{chen2020improved}, we compared the results in Table~\ref{tab:ablation2}. It is clearly seen that CLSA~(68.3\%, 52.5h) can still achieve better results than MoCo V2~(67.5\%, 53h) with the same running time and converge much faster. That also indicates that the optimization of contrastive loss and DDM loss at the same time can also benefit the convergence of representation learning.

\begin{table}[]

\centering
\caption{Comparison of CLSA under different strong augmentations}
\label{tab:strongmag}
\centering
\begin{tabular}{lcccc}
\toprule[1pt]
Strength $s$& 0& 3 & 5 (default) & 7\\ \toprule[1pt]
KNN Acc & 51.9 &52.8 & \textbf{53.0}& 52.6 \\
\bottomrule[1pt]
\end{tabular}
\end{table}

\textbf{The strength of strong augmentation} CLSA achieved promising results by introducing stronger augmentations to further improve self-supervised learning. In contrastive-based approach, the weak augmentations need to be carefully designed to remain the identity. Thus, we designed the stronger augmentations by randomly sampling operations from a large pool of operations with magnitude in a wide range as shown in Table.~\ref{tab:augmentation}. To make the augmentation stronger, we repeated this operation $s$ times. In CLSA, we use $s=5$ as default.  As shown in Table.~\ref{tab:strongmag}, we trained CLSA with 100 epochs and compared the performance with different augmentation strength $s$. Here, the comparison is based on the pre-trained model's KNN classification accuracy, where $K$ is set to 20. CLSA did not influence too much with different augmentation strength for stronger augmentation. However, we can clearly see the improvement compared to the baseline with common weakly augmentation ($s=0$). That suggests the effectiveness and reliability of CLSA framework, which is very independent of the augmentation strength. This property can greatly help CLSA to be adjusted to complement other self-supervised approaches.

\textbf{Representations for weak/strong augmented images} Meanwhile, we also want to compare CLSA with baseline MoCo V2  to illustrate our representative features on both weakly augmented images and strongly images. First, following~\cite{wu2018unsupervised}, we center crop the images to obtain features from the last average pooled layers, and report the accuracy with $K=20$ KNN accuracy. As clearly shown in Fig.~\ref{fig:knn}, on both 200-epoch~(Weak-200) and 800-epoch~(Weak-800) pre-trained encoder, we showed much better performance compared to MoCo V2. Moreover, we further transformed images using our strong augmentations $S$ mentioned in Section \ref{sec:implementation} and obtain the features for them for KNN classification. The comparison is also shown in Fig.~\ref{fig:knn}. It is obvious that we showed much better classification performance on those strongly augmented images. CLSA improved 8.2\% and 9.9\% accuracy compared to MoCo V2 on the 200-epoch~(Strong-200) and 800-epoch~(Strong-800) pre-trained model, respectively. Those comparisons suggest that CLSA can not only increase the representation for weakly augmented images, but also further greatly boost the representation for strongly augmented images. To some extent, we can say the representation learned by CLSA is more general compared to contrastive approaches.

\begin{table}[]

\centering
\caption{Generalization experiments of CLSA}
\label{tab:gen}
\centering
\begin{tabular}{llll}
\toprule[1pt]
Method & Baseline & with CLSA \\ \toprule[1pt]
MoCo V2 (repro)~\cite{chen2020improved} &  67.5 & 69.4 \\
BYOL (repro)~\cite{grill2020bootstrap} & 66.5 &67.3\\  \toprule[1pt]
\end{tabular}
\end{table}

\textbf{Generalization Experiments} To verify the stronger augmentation with DDM loss is beneficial to common contrastive approaches, we further extended our framework to MoCo v2~\cite{chen2020improved} and BYOL~\cite{grill2020bootstrap}. As shown in Table.~\ref{tab:gen}, CLSA showed clear improvement (+1.9\% for MoCo v2; +0.8\% for BYOL) compared to the baseline.  Both experiments are based on asymmetrical loss design for 200 epochs.  Since BYOL is a cosine similarity based method, it's reasonable to see the improvement is relatively small. Hence the experiments verified the generalizability of DDM loss for stronger augmentation for any other contrastive learning based approaches.

\begin{figure}[]
\centering 
\includegraphics[width=0.48\textwidth]{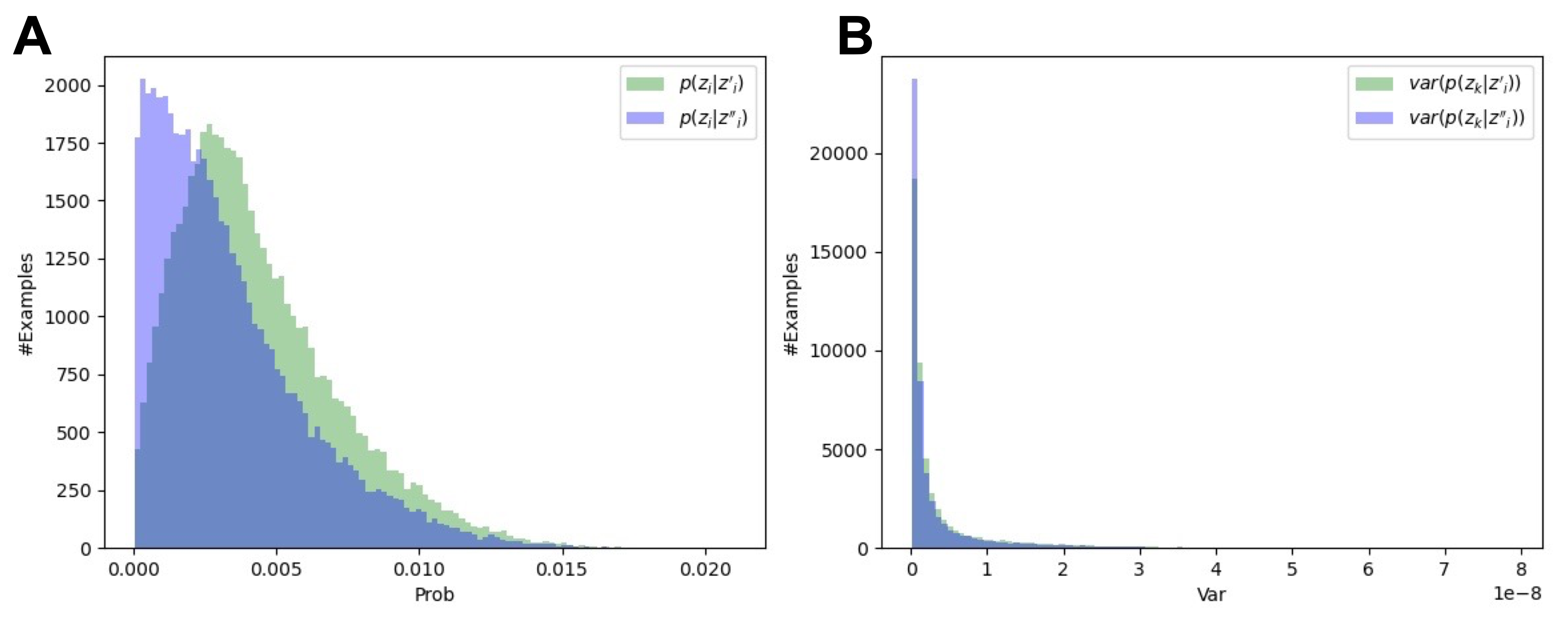} 
\caption{The comparison of the distribution of positive pair's probabilities and variance of negative pairs' probabilities with a pre-trained network by CLSA. \textbf{A.} The distribution comparison of positive pair's probabilities.  \textbf{B.} The distribution comparison of variance of negative pairs' probabilities.} 
\label{fig:clsaweakstrong} 
\end{figure}

\textbf{Distribution of Relative Similarity by CLSA} Similar to the distribution of the probability of positive pairs and the distribution of the variance of the probability of negative pairs shown in Fig.~\ref{fig:clsamotivation}, we further compared that by CLSA in Fig.~\ref{fig:clsaweakstrong}. Compared to contrastive approaches, the distribution of strongly query showed a more similar pattern compared to that of the weakly query. This different distribution clearly suggested that DDM can leverage the knowledge of weakly query to contribute to the learning of strong query. Also, the increase of the mean of the distribution of positive pairs indicates that more and more strongly views can be successfully retrieved by its weakly augmented counterpart. Meanwhile, the increase of the mean of the distribution of the variance of negative pairs also suggests the relationships between the representations of strongly query and other representations is more well captured. In a nutshell, both distribution change agrees well that the feature representation of strongly augmented views become more representative by CLSA pre-training.

\section{Conclusion}
\label{sec:conclusion}
In this paper, we present CLSA, a novel method that can utilize the distributional divergence to learn the information from strongly augmented images. The proposed method outperforms the state-of-the-art methods on all the datasets and achieved almost the same performance compared to the supervised ImageNet network. Meanwhile, it outperforms the previous supervised and self-supervised methods on downstream tasks, which suggests CLSA learned more reliable and fine-grained features that can contribute to the development of other areas. 

Moreover, as aforementioned comparison with contrastive loss, the DDM loss is independent with concurrent contrastive learning methods. For contrastive loss module in CLSA, we can use it to further improve the performance of any contrastive learning methods, such as MoCo~\cite{he2020momentum}, SimCLR~\cite{chen2020simple}, AdCo~\cite{hu2020adco} and so on, where the performance of CLSA can be further boosted with the contrastive baseline BYOL~\cite{grill2020bootstrap}.  Furthermore, the coefficient and the temperature of contrastive loss and distributional divergence minimization loss can be further explored and may benefit more to learn from the strongly augmented images. 

\appendices


\ifCLASSOPTIONcompsoc
  \section*{Acknowledgments}
\else
  \section*{Acknowledgment}
\fi

The authors would like to thank the support of Yuanyuan Zhang for figure plotting. 

\ifCLASSOPTIONcaptionsoff
  \newpage
\fi



\bibliographystyle{IEEEtran}
\bibliography{reference}

%

%



\begin{IEEEbiography}[{\includegraphics[width=1in,height=1.25in,clip,keepaspectratio]{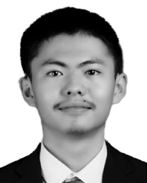}}]{Xiao Wang}
Xiao Wang received his B.S. degree from Department of Computer Science in Xi’an Jiaotong University, Xi’an, China in 2018. He is currently pursuing his Ph.D. degree in the Department of Computer Science at Purdue University, West Lafayette, IN, USA. His research interests include deep learning, computer vision, bioinformatics and intelligent systems. 
\end{IEEEbiography}

\begin{IEEEbiography}[{\includegraphics[width=1in,height=1.25in,clip,keepaspectratio]{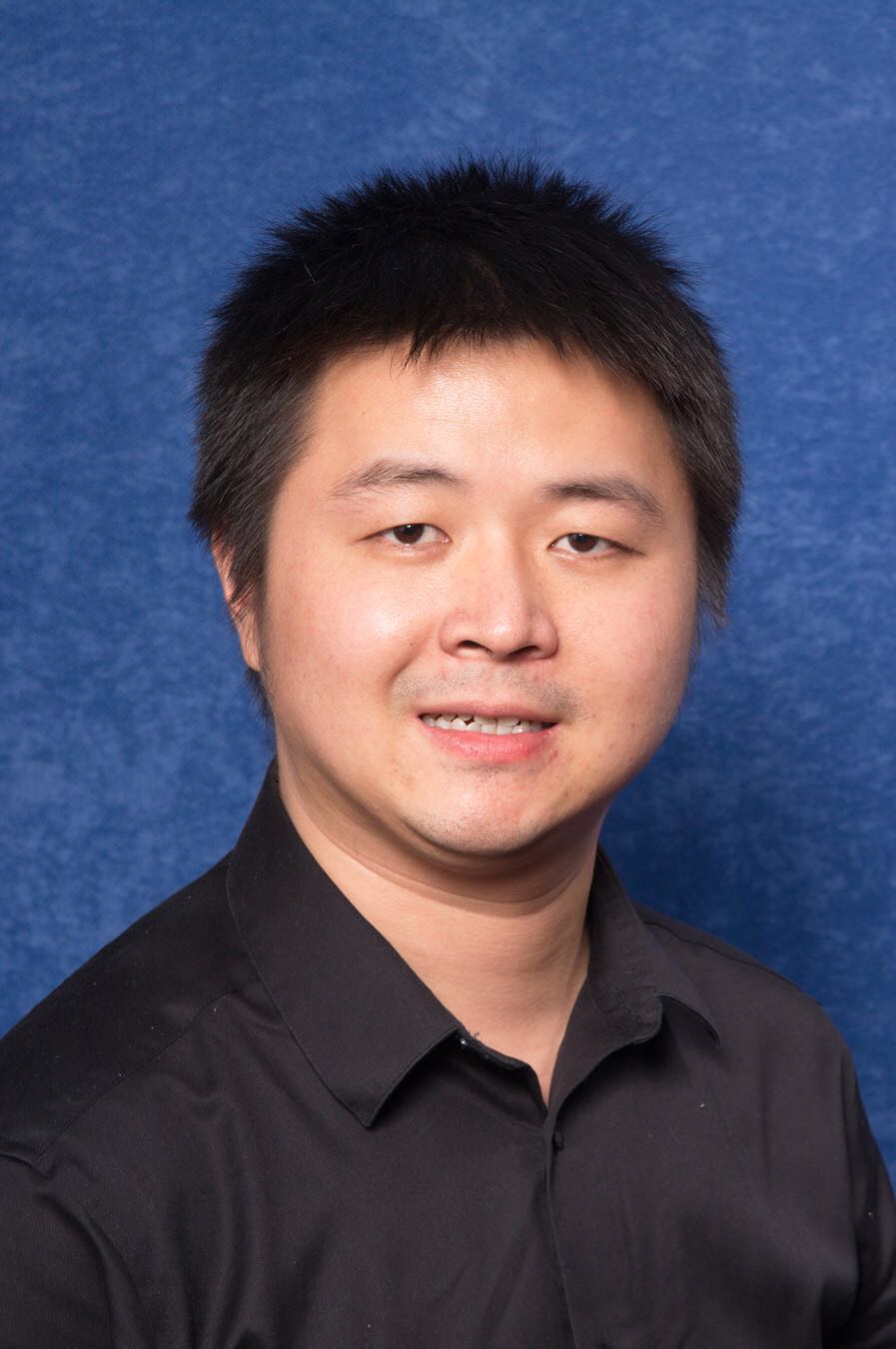}}]{Guo-Jun Qi}
Guo-Jun Qi (M14-SM18) is the
Chief Scientist leading and overseeing an international R\&D team for multiple artificial intelligent
services on the Huawei Cloud since August
2018. He was a faculty member in the Department
of Computer Science and the director of
MAchine Perception and LEarning (MAPLE) Lab
at the University of Central Florida since August
2014. Prior to that, he was also a Research Staff
Member at IBM T.J. Watson Research Center,
Yorktown Heights, NY. His research interests include
machine learning and knowledge discovery from multi-modal data
sources to build smart and reliable information and decision-making
systems. Dr. Qi has published more than 100 papers in a broad range
of venues in pattern recognition, machine learning and computer vision.
He also has served or will serve as a general co-chair for ICME 2021,
technical program co-chair for ACM Multimedia 2020, ICIMCS 2018
and MMM 2016, as well as an area chair (senior program committee
member) for multiple academic conferences. Dr. Qi is an associate editor
for IEEE Transactions on Circuits and Systems for Video Technology (TCSVT),
IEEE Transactions on Multimedia (T-MM), IEEE Transactions on
Image Processing (T-IP), Pattern Recognition (PR), and ACM Transactions
on Knowledge Discovery from Data (T-KDD).
\end{IEEEbiography}

\end{document}